%% file: main.tex
\documentclass{article}

\PassOptionsToPackage{numbers}{natbib}

\usepackage[preprint]{neurips_2025}

\usepackage[utf8]{inputenc}
\usepackage[T1]{fontenc}
\usepackage{xcolor}

\usepackage{hyperref}
\hypersetup{
    colorlinks=true,
    linkcolor=blue!55!black,  
    citecolor=blue!55!black,  
    urlcolor=blue!55!black    
}
\usepackage{url}
\usepackage{booktabs}
\usepackage{amsfonts}
\usepackage{amsmath}
\usepackage{amssymb}
\usepackage{nicefrac}
\usepackage{microtype}
\usepackage{wrapfig}
\usepackage{enumitem}
\usepackage{graphicx}
\usepackage{float}
\usepackage{adjustbox}
\usepackage{multirow}
\usepackage{array}
\usepackage{algorithm}
\usepackage{algpseudocode}
\usepackage{pifont}
\usepackage{tikz}
\usetikzlibrary{positioning, arrows.meta}

\usepackage{booktabs}
\usepackage{tabularx}
\usepackage{multirow} 
\usepackage{array}

\usepackage{algorithm}
\usepackage{algpseudocode}

\title{Spark-to-Paper: End-to-End Research Paper Generation as a Composable Skill}

\author{
Zhuoyang Qian$^{1,*}$, Biao Wu$^{2,*}$, Yiran Wang$^{1,*}$,
Chris D Yan$^{1,*}$, \\
\textbf{Desan Dai}$^{1}$, \textbf{Liangwei Zheng}$^{1}$, \textbf{Jin Jiang}$^{1}$, \textbf{Jusheng Zhang}$^{1}$,
\textbf{Wenhao Wang}$^{1,\ddagger}$ \\
$^{1}$Vast Intelligence Lab 
$^{2}$University of Technology Sydney\\
$^{*}$Equal Contribution 
$^{\ddagger}$Corresponding Author\\
\texttt{https://github.com/Spark-To-Paper-Skills/spark-to-paper-skills}\\
\texttt{wangwenhao@vastilab.com}
}

\begin{document}
\maketitle
\begin{abstract}
Turning a research idea into a complete paper requires more than text generation: the system must retrieve literature, design and execute experiments, revise claims according to evidence, produce publication-ready figures, and maintain consistency across a long generation process. We present \textbf{Spark-to-Paper}, an end-to-end research paper generation system implemented as thirteen composable skills inside an existing coding assistant, without requiring a separate agent platform or orchestration service. Spark-to-Paper separates model-based judgment from deterministic operations that can be directly executed and checked. It further separates experiment planning from reporting, so that required evidence is specified before results are observed and manuscript claims are revised according to measured outcomes. To improve reliability over long research trajectories, the system combines deterministic integrity checks with self-critique and bounds a failure mode we call the Self-Refutation Loop, in which repeated experiments continue to reject the original research objective. Spark-to-Paper also produces editable vector figures through programmatic plotting for experimental results and code-based reconstruction for generated method diagrams. Across eight controlled research topics, Spark-to-Paper achieves 99.5\% citation validity and 96.4\% figure editability. A controlled ablation increases fabrication detection from 14\% for a single-pass draft to 92\% with the full integrity and review stack, while adversarial review achieves 74\% precision. The full system uses 11.9M tokens, costs \$8.1 per manuscript, and requires 3.2 hours on average. These results show that end-to-end research paper generation can be implemented as a lightweight, composable workflow inside existing coding assistants while keeping experimental evidence central to how claims are accepted, revised, or abandoned.
\end{abstract}

\input{sections/introduction.tex}
\input{sections/related_work.tex}

\input{sections/approach.tex}

\input{sections/pipeline.tex}
\input{sections/integrity.tex}

\input{sections/figure_engine.tex}
\input{sections/evaluation.tex}

\input{sections/discussion.tex}
\input{sections/conclusion.tex}

\newpage 

\bibliographystyle{plainnat}
\bibliography{refs}

\appendix

\input{sections/appendix}

\end{document}

%% file: sections/introduction.tex
\section{Introduction}

Turning a research idea into a complete academic paper requires substantially more than generating text. A researcher must identify relevant literature, design and run experiments, decide whether the resulting evidence supports the original hypothesis, revise claims when it does not, produce publication-ready figures, and maintain consistency across a manuscript that evolves over many stages. Recent autonomous research agents have begun to automate this broader process, demonstrating that language models can participate in ideation, experimentation, review, and paper generation~\cite{lu2024aiscientist,yamada2025aiscientistv2,schmidgall2025agentlab,ghareeb2025robin,xu2026idea2story}. However, these systems are typically implemented as standalone applications with their own orchestration layers and supporting infrastructure. This makes them powerful, but also separate from the coding environments in which much of the actual research work already takes place.

At the same time, modern coding assistants already provide many of the basic capabilities required for research automation: they can inspect project files, execute code, search for information, call external tools, and revise artifacts over long interactions~\cite{anthropic2026claudecode,anthropic2026agentsdk}. Existing skill-based research tools exploit some of these capabilities for tasks such as literature search, outlining, and manuscript drafting~\cite{wu2026ars}, but generally stop short of the complete research-to-paper process. This raises a simple question: \textit{can end-to-end research paper generation be implemented as a collection of reusable skills inside an existing coding assistant, rather than as a separate autonomous research platform?}

We present \textbf{Spark-to-Paper}, a system that realizes this design through thirteen composable skills. Each skill handles a research task, such as planning, literature retrieval, writing, review, figure generation, or experiment execution, while all skills communicate through artifacts in a shared project directory. The coding assistant determines how each skill should be executed from the current project state, while a lightweight pipeline specifies only the high-level order of tasks. Within each skill, we further separate judgment from execution: the language model handles decisions that require interpretation or reasoning, whereas deterministic programs handle operations that can be explicitly executed and checked.

Three components are particularly important for extending this design beyond ordinary paper drafting. \textbf{First}, Spark-to-Paper separates experiment planning from experiment reporting. The evidence required by the manuscript is specified before results are observed, experiments are executed to resolve missing evidence, and the resulting artifacts are used to revise claims throughout the paper. \textbf{Second}, the system combines deterministic integrity checks with long-horizon self-critique~\cite{shinn2023reflexion,madaan2023selfrefine}. We additionally identify a failure mode that we call a \textbf{Self-Refutation Loop}, in which the system repeatedly concludes that its own experiments fail to support the original research objective and continues revising the same direction. Spark-to-Paper bounds this process, records unsuccessful trajectories as failure reports, and restarts from a new idea rather than forcing every research direction to succeed. \textbf{Third}, the figure pipeline separates quantitative plots from explanatory figures. Experimental plots are generated directly from measured data, while method figures are first designed with an image-generation model and then reconstructed through code into editable vector PDFs.

We evaluate Spark-to-Paper through controlled experiments, retrospective audits, and qualitative case studies. Across eight controlled research topics, the full system achieves 99.5\% citation validity and 96.4\% figure editability. In a controlled ablation with injected unsupported claims, fabrication detection increases from 14\% for a single-pass draft to 92\% with the complete integrity and review stack, while adversarial review achieves 74\% precision on raised issues. The full system uses 11.9M tokens on average, costs \$8.1 per manuscript, and requires 3.2 hours of wall-clock time in our instrumented runs. We report these measurements together with audits and disclosed costs from prior systems, while avoiding direct price equivalence where the underlying workloads differ.

Our contributions are threefold:

\begin{itemize}[leftmargin=*, itemsep=0pt]

\item We introduce Spark-to-Paper, an end-to-end research paper generation system implemented as thirteen composable skills inside an existing coding assistant, without requiring a separate agent platform or orchestration service.

\item We develop mechanisms for evidence-grounded long-horizon research generation, including pre-committed experiment design, evidence-guided claim revision, deterministic integrity checks, bounded recovery from self-refutation loops, and code-reconstructed editable figures.

\item We provide a controlled evaluation of the resulting research artifacts, including citation validity, fabrication detection, figure editability, review precision, robustness, and generation cost, together with ablations and qualitative analyses of system behavior and failure modes.

\end{itemize}

%% file: sections/related_work.tex
\section{Related Work}
\label{sec:related_work}

\begin{table}[t]
\centering
\small
\caption{Qualitative capability comparison against directly related systems, based on each system's own public documentation (paper or repository) as of this writing. \ding{51}\ full, \ding{108}\ partial, --\ not offered or not documented.}
\label{tab:comparison}
\resizebox{\linewidth}{!}{%
\begin{tabular}{lccccc}
\toprule
System & End-to-end & Runs exper.\ & Draws figures & Editable vectors & No standing infra.\ \\
\midrule
AI Scientist / v2 \citep{lu2024aiscientist,yamada2025aiscientistv2} & \ding{51} & \ding{51} & \ding{51} & -- & -- \\
AutoResearchClaw \citep{liu2026autoresearchclaw} & \ding{51} & \ding{51} & \ding{51} & -- & -- \\
Kosmos / Robin \citep{mitchener2025kosmos,ghareeb2025robin} & \ding{108} & \ding{51} & \ding{108} & -- & -- \\
Idea2Story \citep{xu2026idea2story} & -- & -- & \ding{108} & -- & \ding{51} \\
ARS \citep{wu2026ars} & \ding{108} & -- & -- & -- & \ding{51} \\
CycleResearcher \citep{weng2025cycleresearcher} & \ding{108} & \ding{51} & -- & -- & -- \\
\textbf{Spark-to-Paper (ours)} & \ding{51} & \ding{51} & \ding{51} & \ding{51} & \ding{51} \\
\bottomrule
\end{tabular}%
}
\vspace{-3mm}
\end{table}

\subsection{Autonomous End-to-End Research Agents}

A growing line of work runs an entire research cycle, \textit{i.e.}, ideation, experimentation, writing, and often self-review, as a standalone autonomous system. \citet{lu2024aiscientist} and its successor \citet{yamada2025aiscientistv2} generate, run, and write up machine learning experiments with a simulated review stage; \citet{liu2026autoresearchclaw} extends this to a self-reinforcing, human-in-the-loop pipeline reported at a scale of tens of thousands of lines of code; \citet{mitchener2025kosmos} and \citet{ghareeb2025robin} target open-ended scientific discovery over long, multi-hour autonomous runs with independently-audited statement accuracy; 
\begin{wrapfigure}{r}{0.55\linewidth}
    \centering
    \vspace{-3mm}
    \includegraphics[width=\linewidth]{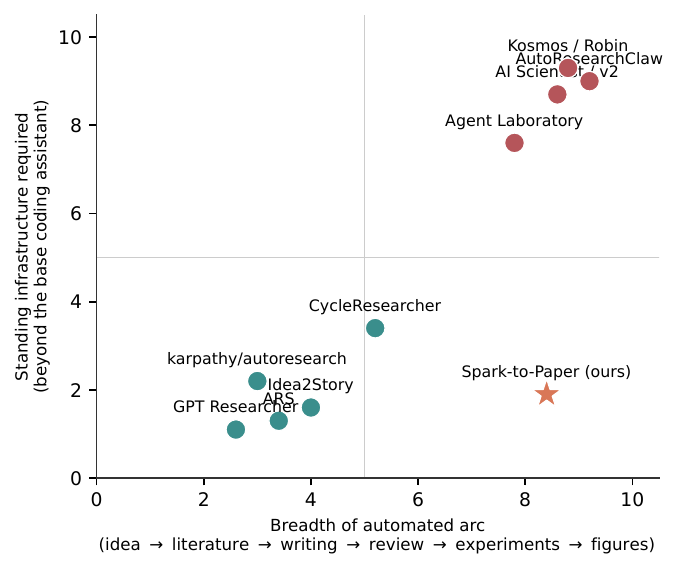}
    \vspace{-6mm}
    \caption{Qualitative positioning against directly related systems, on two axes derived from Table~\ref{tab:comparison}. Coordinates are the authors' own ordinal assessment from each system's public documentation, not a measured benchmark.}
    \label{fig:positioning}
    \vspace{-9mm}
\end{wrapfigure}
\citet{deepmind2026coscientist} and \citet{schmidgall2025agentlab} similarly pair an LLM-driven experimentation loop with a writing stage. These systems match or exceed Spark-to-Paper's breadth, particularly in autonomously proposing and running novel experiments, but each ships as a standalone application, typically a dedicated codebase with its own orchestration server, and in several cases a graph database or cluster scheduler, rather than as skills that run inside an existing coding assistant with no additional service to operate.

\subsection{Lightweight, Skill-Based Writing Assistants}

A second line of work stays inside a coding assistant's existing skill or plugin mechanism rather than shipping a separate application. \citet{xu2026idea2story} and \citet{wu2026ars} decompose literature search, drafting, and review into composable steps much as Spark-to-Paper does; \citet{elovic2026gptresearcher} and \citet{karpathy2026autoresearch} automate a narrower slice, \textit{i.e.}, web-grounded report writing, and single-GPU experiment iteration, respectively; \citet{chen2026autoresearchsurvey} surveys this emerging category directly. \citet{weng2025cycleresearcher} pairs an automated researcher with an automated reviewer trained specifically to critique it, the closest existing analogue to Spark-to-Paper's own adversarial review stage. None of these tools, however, runs real experiments end to end or produces editable vector figures rather than embedded bitmaps, the two capabilities at the center of this paper's contributions.

\subsection{LLM Agents, Tool Use, and Self-Critique}

Spark-to-Paper's internal mechanisms draw on, without directly reimplementing, a broader agent literature. Its reasoning-then-acting stage structure follows the interleaved reasoning-and-action pattern of \citet{yao2023react}; its use of external tools (web search, a DOI resolver, a plotting library) for tasks a language model should not do itself echoes \citet{schick2023toolformer} and \citet{nakano2021webgpt}; its refine and review stages apply a model's own critique of its output in the manner of \citet{shinn2023reflexion} and \citet{madaan2023selfrefine}; its planning stage decomposes a paper into an outline before any prose is written, in the same spirit as deliberate multi-step reasoning approaches \citep{wei2022chainofthought, yao2023tot}; and its persistent, file-backed working state across skill invocations resembles the memory-and-reflection architecture of \citet{park2023generativeagents}, adapted here to a single long-running document rather than a simulated society of agents.

\subsection{Grounded Generation and Citation Integrity}

The concern motivating Spark-to-Paper's gate infrastructure is well documented: large language models fabricate citations and factual claims at measurable, non-trivial rates when asked to write about literature they have not actually retrieved \citep{walters2023fabrication, ji2023hallucination, zhang2023sirensong}. Retrieval-augmented generation is the standard mitigation \citep{lewis2020rag, shuster2021retrieval, gao2023ragsurvey}, and Spark-to-Paper's citation stage is best understood as a retrieval-then-verify instance of this pattern, specialized to bibliographic metadata rather than passage-level facts. Its no-fabrication rule for tabular results connects to a related but distinct literature on generating text from structured data faithfully \citep{moosavi2021scigen}, where the risk is not an invented source but an invented number.
Taken together, this body of work establishes both ends of the space Spark-to-Paper occupies: systems that automate more of the research process but require standing infrastructure to do it, and systems that stay infrastructure-free but automate a narrower slice of it. Positioning a new system between two established poles is easy to assert and harder to substantiate, since it invites exactly the kind of overclaim this paper otherwise argues against; 
Table \ref{tab:comparison} and Fig. \ref{fig:positioning} are our attempt to make that positioning checkable against each compared system’s own public documentation rather than leaving it as a rhetorical claim. The following sections substantiate this positioning in mechanistic detail and, in Sections \ref{sec:integrity} and \ref{sec:evaluation}, examine the trade-offs that accompany this design.


%% file: sections/approach.tex
\section{Spark-to-Paper}
\label{sec:spark_to_paper}

Spark-to-Paper is a collection of thirteen skills that runs inside an existing
coding assistant. Each skill handles a specific research task, such as planning
a paper, searching for literature, writing and revising the manuscript,
reviewing claims, generating figures, or running experiments. Rather than
building a separate agent system, Spark-to-Paper uses the coding assistant's
existing abilities to read files, execute code, search for information, and
call external tools. The skills share a common project directory. Each skill reads the artifacts
produced so far, performs its task, and writes its output back to the project.
For example, the planning skill produces a paper blueprint, the citation skill
builds a verified bibliography, and the writing skill uses both to generate
the manuscript. Later skills operate on the same manuscript, figures, and
experimental outputs. The project files therefore provide a simple interface
between skills and allow the paper to evolve across the full pipeline. In our implementation, Spark-to-Paper runs inside Claude Code using a model
from the Claude family
\citep{anthropic2024claude3, anthropic2026claudecode,
anthropic2026agentsdk}. The design itself is not tied to Claude Code. A coding
assistant with comparable abilities to inspect project files, use tools, and
execute code can support the same skill-based design.

\textbf{Skills as the execution unit.}
A skill defines what a research task should accomplish, the constraints it must satisfy, the tools it may use, and the artifacts it should produce. It does not specify every reasoning step needed to complete the task. Instead, the coding assistant decides how to carry it out from the current project state. For example, the refinement skill may only reorganize an argument in one paper, while another may require additional citations or changes after a failed consistency check. Both use the same skill. This differs from encoding the full system behavior as a fine-grained agent graph, where individual actions and transitions must be specified in advance. Spark-to-Paper nevertheless maintains a simple high-level order for paper generation. The pipeline specifies \emph{which task comes next}, while each skill determines \emph{how that task is completed}. This keeps the process easy to follow without fixing the detailed execution path of every research task.

\textbf{Model and deterministic tools.}
Within each skill, we separate tasks that require judgment from tasks that can
be executed and checked mechanically. The language model handles decisions
such as how to organize an argument, which literature is relevant, whether
available evidence supports a claim, and how a result should be presented.
These decisions depend on the manuscript and the evidence available at that
point in the project. Operations with explicit rules are handled by deterministic scripts. Examples
include checking manuscript structure, validating citations, compiling
\LaTeX, plotting measured results, and checking generated files. These
operations do not benefit from open-ended generation and can instead be
executed reproducibly and verified directly. The resulting design follows a
simple division of responsibility: the model handles judgment, while code
handles operations that can be explicitly executed and checked.

\subsection{From Input to Paper}

\begin{figure}[!t]
\centering
\includegraphics[width=\linewidth]{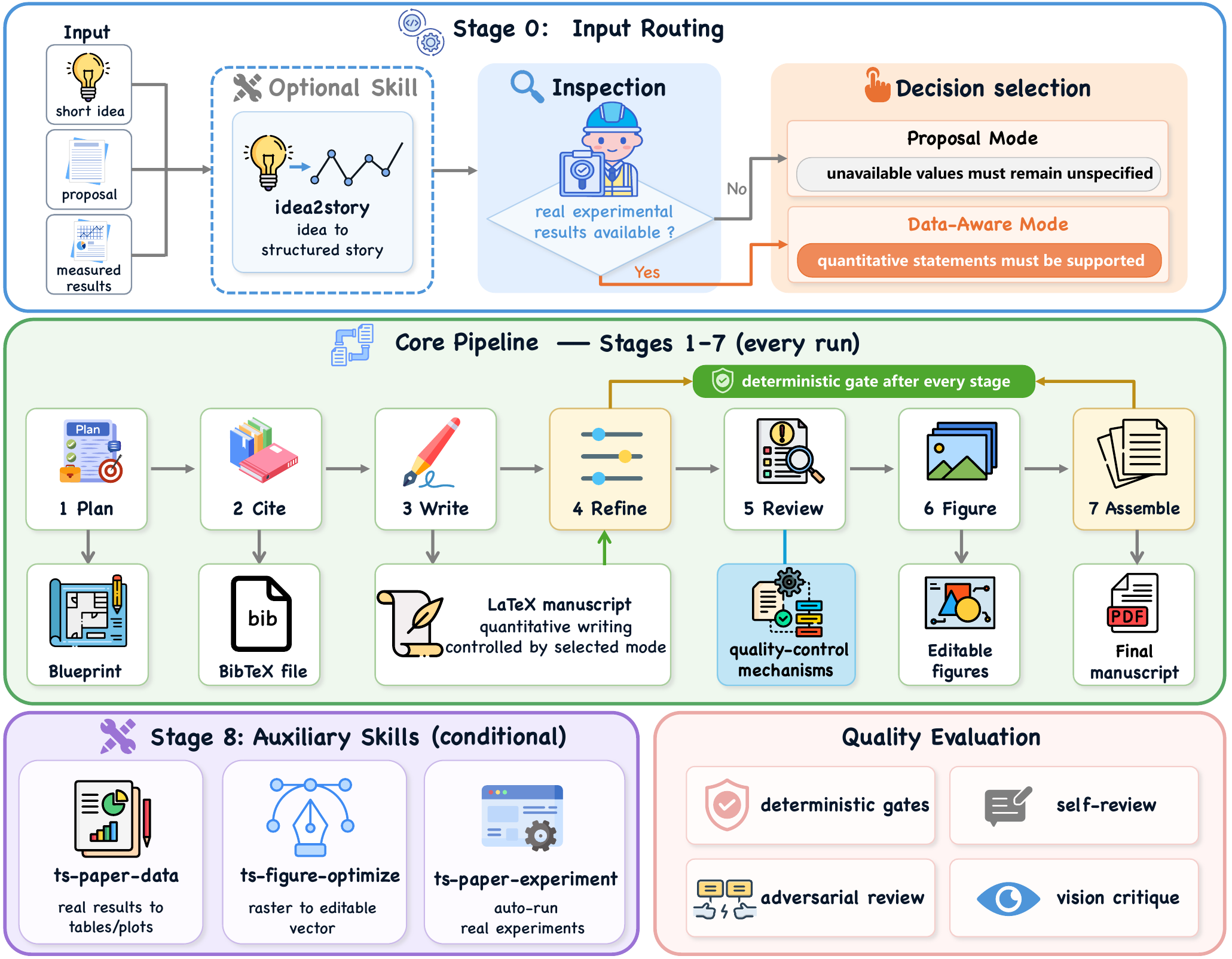}
\vspace{-4mm}
\caption{Overview of Spark-to-Paper execution. Stage~0 selects the result-integrity mode, Stages~1--7 coordinate through persistent project artifacts, and conditional Stage~8 writes measured evidence back into the manuscript. Model judgment governs context-dependent decisions, while deterministic tools execute checkable operations.}
\vspace{-2mm}
\label{fig:system_execution}
\vspace{-3mm}
\end{figure}

As shown in Fig. \ref{fig:system_execution}, the thirteen skills provide the reusable building blocks of Spark-to-Paper.
Their execution is organized by the \texttt{ts-paper} orchestrator into an
input-routing step, seven core paper-generation stages, and an experiment
stage. Additional skills support particular inputs or stages, such as
expanding a short research idea, processing measured results, or converting
generated figures into editable formats. Thus, the number of skills does not correspond one-to-one with pipeline stages.

\noindent
\textbf{Stage 0: Input Routing.}
Spark-to-Paper first inspects the user input and determines where the pipeline
should begin. A short research idea can first be expanded into a structured
research proposal, whereas an already developed proposal can enter the paper
pipeline directly. The system also determines whether real experimental
results are available. This decision selects one of two result-integrity modes. In
\textbf{Proposal Mode}, the input contains no measured results. The system may
design experiments and prepare result tables, but unavailable values must
remain unspecified. In \textbf{Data-Aware Mode}, measured data or experimental
outputs are available, and quantitative statements in the manuscript must be
supported by those results. This mode is propagated through the remaining
stages so that the same generation pipeline can operate both before and after
experiments have been conducted.

\noindent
\textbf{Stage 1: Planning.}
The planning skill converts the input into a structured paper blueprint. It
identifies the research question, main contributions, section structure,
notation, and experimental design. It also reads the specification of the
target venue to determine structural requirements such as section organization
and length constraints. The resulting blueprint becomes the main specification
used by later stages.

\noindent
\textbf{Stage 2: Citation.}
The citation skill builds the bibliography required by the planned paper. It
searches for relevant literature and verifies candidate references using
bibliographic information such as DOI records, arXiv identifiers, and other
metadata. The goal is to find verifiable support for the claims and context
required by the blueprint rather than to generate references from model
memory. Verified references are stored in a BibTeX file and reused throughout
the manuscript.

\noindent
\textbf{Stage 3: Writing.}
The writing skill generates the complete \LaTeX{} manuscript from the
blueprint and verified bibliography. All sections are written against the same
project context so that terminology, notation, contributions, and experimental
claims remain consistent across the paper. The selected integrity mode also
controls quantitative writing: Proposal Mode leaves unavailable results
unspecified, while Data-Aware Mode allows measured values supported by the
provided data.

\noindent
\textbf{Stage 4: Refinement.}
The refinement skill revises the manuscript as a whole rather than treating
sections independently. It removes repetition, reconciles terminology and
notation, improves local arguments, and adjusts section lengths to the target
template. Deterministic checks are then rerun to ensure that revision has not
introduced structural, citation, or integrity errors.

\noindent
\textbf{Stage 5: Review.}
The review skill challenges the current manuscript before it is finalized.
Multiple isolated review passes examine complementary aspects of the paper,
such as technical soundness, experimental design, and the strength of the
evidence. Review comments are tied to specific passages in the manuscript and
are checked before being accepted as valid issues. Confirmed problems are
returned to the refinement stage for correction. We describe the corresponding
quality-control mechanisms in Section~\ref{sec:integrity}.

\noindent
\textbf{Stage 6: Figure Generation.}
The figure skill generates figures according to their role in the paper.
Plots that represent measured results are produced programmatically from the
underlying data. Method diagrams, system architectures, and other explanatory
figures may instead be produced with an image-generation model. Generated
raster figures are subsequently processed into editable artifacts when
possible, while preserving their approved visual content. The figure pipeline
is described in detail in Section~\ref{sec:figure_engine}.

\noindent
\textbf{Stage 7: Assembly.}
The assembly skill combines the manuscript sections, bibliography, figures,
and venue template into a complete \LaTeX{} project. Template-specific
operations such as section ordering, citation formatting, and document
structure are derived from the template specification rather than hard-coded
for a single venue. The project is then compiled, and deterministic checks are
used to verify that the final manuscript contains no unresolved citations or
\LaTeX{} errors.

\noindent
\textbf{Stage 8: Experiment Execution.}
Once a complete first draft is available, Spark-to-Paper checks whether the
planned experiments can be executed using the available code and data. When
they are feasible, the experiment skill runs them, records the measured
outputs, and uses the results to update the manuscript. Result tables and
data-dependent figures are regenerated from the measurements, and claims in
the abstract, introduction, experiments, and conclusion can be revised to
match the new evidence. If the required code or data are unavailable, the system leaves the corresponding results unspecified rather than generating values. The
experiment stage therefore connects the proposal and data-aware parts of the
pipeline: the earlier stages specify what evidence the paper needs, while this
stage obtains that evidence when the required resources are available. We describe experiment execution in detail in Appendix~\ref{app:experiment_execution}.

\subsection{Key Components}
\label{sec:key_components}

The pipeline above summarizes the overall execution of Spark-to-Paper. We leave most implementation details, including individual skill specifications, intermediate artifacts, and auxiliary checks, to Appendices~\ref{app:implementation} and~\ref{app:deterministic_gates}. In the main text, we focus on three components that are particularly important for end-to-end paper generation: generating publication-ready and editable figures, planning and executing experiments based on the evidence required by the manuscript, and maintaining reliable reasoning over long generation horizons through repeated self-critique and claim revision. We describe these components in the following sections.


%% file: sections/pipeline.tex
\section{Experiment Design and Revision}
\label{sec:experiments-stage}

Spark-to-Paper separates experiment planning from experiment reporting to reduce the risk of adapting the evaluation protocol after observing the results. Before execution, the planning stage specifies the datasets, baselines, metrics, ablations, and result tables required to evaluate the manuscript's main claims. The table structure is fixed in advance, while numerical cells remain empty until the corresponding experiments are completed. This creates a lightweight form of preregistration within the pipeline: the system decides what evidence would be required before seeing that evidence, and the experiment stage inherits this design rather than redefining it based on observed outcomes.

The experiment stage then maps manuscript claims to the evidence needed to support them and identifies which experiments are still missing. Rather than exhaustively running every experiment suggested by a generic template, Spark-to-Paper executes the minimal set needed to resolve these evidence gaps. Paper-type-specific guidance may recommend suitable baselines, metrics, ablations, and common failure checks, but does not override the committed design. Expensive runs or external data acquisition must be justified by the specific claim they support and by the availability of the required resources. Experimental execution produces traceable artifacts, including logs, metric files, tables, and figures, rather than manuscript claims directly. A numerical result is allowed to enter the paper only when it can be traced back to its dataset, model configuration, seed, metric, and source output.

After execution, Spark-to-Paper revisits the manuscript according to the resulting evidence. Claims are classified as \textsc{supported}, \textsc{partially-supported}, \textsc{unsupported}, \textsc{contradicted}, or \textsc{needs-confirmation}, and are retained, weakened, removed, moved to limitations, or used to trigger additional experiments accordingly. Null, negative, or inconclusive results are preserved rather than omitted, and the resulting changes are propagated across the manuscript so that the abstract, introduction, results, and conclusion remain consistent with the final evidence. Design commitment and several artifact-level checks are machine-checked, while claim-level evidence diagnosis is currently performed by the model and recorded in structured reports. We provide the full claim admission protocol and additional implementation details in Appendices~\ref{app:claim_admission} and~\ref{app:implementation}.

%% file: sections/integrity.tex
\section{Integrity and Self-Critique}
\label{sec:integrity}

Long-horizon paper generation introduces two different types of errors. Some are explicit and machine-checkable, such as broken citations, missing figures, compilation failures, or numerical results appearing without supporting data. Others are semantic and emerge as the manuscript evolves: an early claim may become inconsistent with later evidence, an argument may drift across sections, or a decision made earlier in the pipeline may no longer remain appropriate. Spark-to-Paper handles these two cases differently. Deterministic checks enforce constraints that can be stated explicitly, while model-based self-critique revisits decisions that require semantic judgment. Together, they provide hard checks for verifiable properties and iterative correction for errors that emerge over long generation horizons.

\begin{figure}[!t]
\centering
\includegraphics[width=\linewidth]{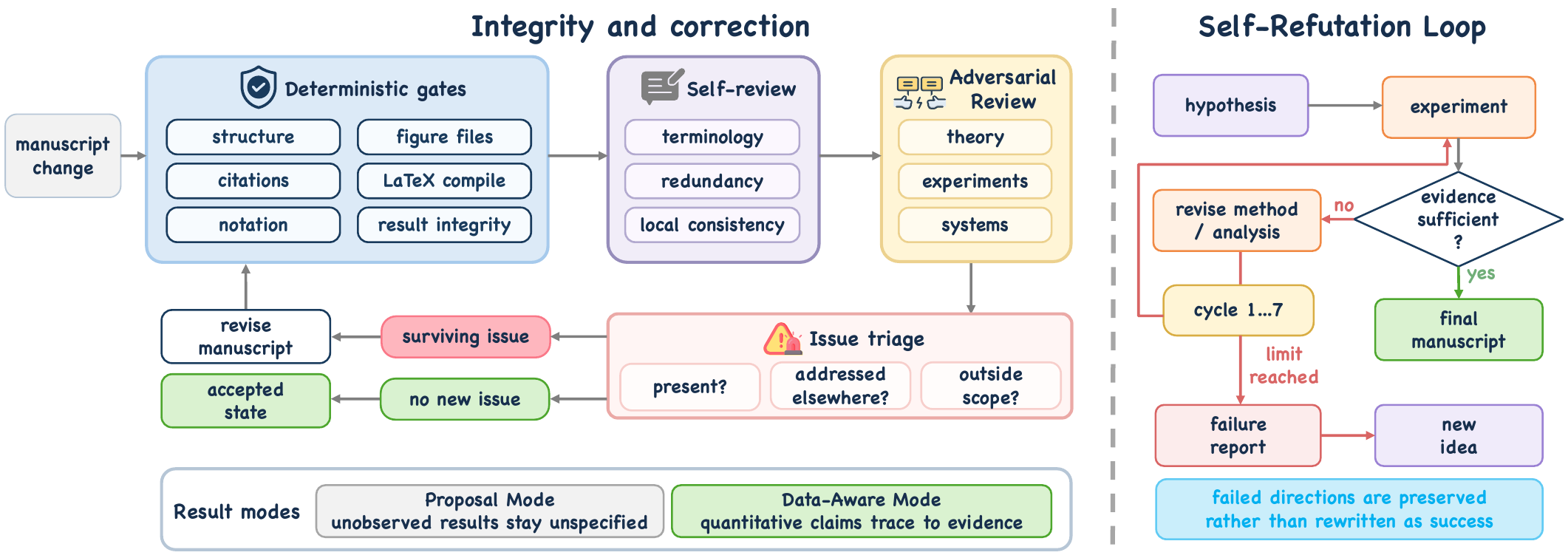}
\vspace{-5mm}
\caption{Integrity and correction in Spark-to-Paper. Deterministic gates enforce verifiable properties, Self-Review and Adversarial Review challenge semantic decisions, and surviving issues trigger revision. Experiment--critique--revision cycles are capped at seven; an unresolved trajectory becomes a failure report rather than a successful manuscript.}
\label{fig:integrity_loops}
\vspace{-4mm}
\end{figure}

\subsection{Deterministic Integrity Checks}

For properties with explicit correctness criteria, Spark-to-Paper uses deterministic gates. The current implementation checks the paper blueprint, citations, manuscript content, figures, and the compiled \LaTeX{} project, covering requirements such as template structure, citation consistency, notation, required figure files, and successful compilation. The same mechanism also enforces result integrity. In \textbf{Proposal Mode}, experimental results that have not been observed must remain unspecified rather than being filled with generated values; in \textbf{Data-Aware Mode}, quantitative statements must be supported by provided data or experimental outputs. A stage proceeds only after its applicable checks pass. Because these conditions can be stated and verified directly, they are implemented as deterministic programs rather than model judgments. Detailed gate definitions and failure conditions are provided in Appendix~\ref{app:deterministic_gates}.

\subsection{Self-Critique over Long-Horizon Generation}

Deterministic checks cannot determine whether an argument is sufficiently supported, whether different sections remain conceptually consistent, or whether an earlier claim should be revised after new evidence becomes available. Spark-to-Paper therefore uses two levels of model-based critique. \textbf{Self-Review} operates locally after an edit, checking the modified content against the surrounding manuscript and repairing terminology drift, redundancy, and local inconsistencies. \textbf{Adversarial-Review} operates at the manuscript level, where multiple isolated review passes examine the paper from complementary perspectives, including theoretical soundness, experimental design, and systems validity. Every proposed issue must identify and quote the specific passage it challenges, and is then checked from three directions: whether the problem is actually present, whether it has already been addressed elsewhere in the manuscript, and whether it falls outside the stated scope of the work. Issues that can be refuted are discarded, while those that survive are returned for revision. Review continues until an additional round produces no new surviving issues.

Together, these steps form a long-horizon correction loop: the system makes a claim or writing decision, later challenges it, attempts to refute the challenge, and revises the manuscript when the challenge remains valid. This allows later evidence to change earlier decisions rather than simply being appended to them. For example, experimental results may weaken an earlier claim and trigger revisions to the experiments section, introduction, abstract, or conclusion. A similar principle is used for figures, where reference grounding and visual review compare generated diagrams against the manuscript and revise them when necessary (Section~\ref{sec:figure_engine}).

\subsection{Self-Refutation Loops and Bounded Recovery}

A failure mode we observe in long-horizon research generation is a \textbf{Self-Refutation Loop}. The system begins with a research hypothesis or design objective, plans experiments to test it, and then evaluates whether the observed results provide sufficient evidence for the intended claim. When the evidence is judged insufficient, the model may revise the method, experimental design, or analysis and run another round of experiments. In some cases, however, the new evidence is again judged insufficient, causing the system to repeat the same cycle of experimentation, critique, and revision. This behavior is different from ordinary iterative improvement. A productive iteration resolves a concrete weakness or produces new evidence that moves the research toward a clearer conclusion. In a self-refutation loop, the system repeatedly concludes that its own results do not adequately support the original hypothesis or design objective, yet continues revising the same research direction rather than accepting that the idea may simply not work. The manuscript and experiments may therefore continue to change without converging to a defensible scientific conclusion.

Spark-to-Paper does not assume that such loops can always be resolved. Instead, we bound the number of experiment--critique--revision cycles to seven. If the research objective still cannot be supported after this limit, the current trajectory is terminated. The system records the original idea, the attempted methods and experiments, the observed results, and the reasons why the evidence was insufficient in a failure report. This report is treated as the outcome of an unsuccessful research direction rather than being rewritten into an apparently successful paper. The system then starts a new research trajectory from a different idea and reruns the full pipeline. Some trajectories may fail in the same way, while others may produce experimental evidence that supports a coherent set of claims. Only the latter are consolidated into a final manuscript. In this way, Spark-to-Paper does not force every research idea to succeed: it bounds self-refutation loops, preserves failed attempts as research outcomes, and moves on to alternative ideas when the current hypothesis cannot be supported.

%% file: sections/figure_engine.tex
\section{Editable Figure Generation}
\label{sec:figure_engine}

Spark-to-Paper generates paper figures through two different paths depending on their role. Method and explanatory figures are first created visually with an image-generation model and then reconstructed as editable vector graphics through code. Figures that report experimental results are instead generated directly from measured data using plotting programs. This separation allows the system to use generative models for visual design while keeping quantitative figures strictly grounded in experimental results.

\begin{figure}[!t]
\centering
\includegraphics[width=\linewidth]{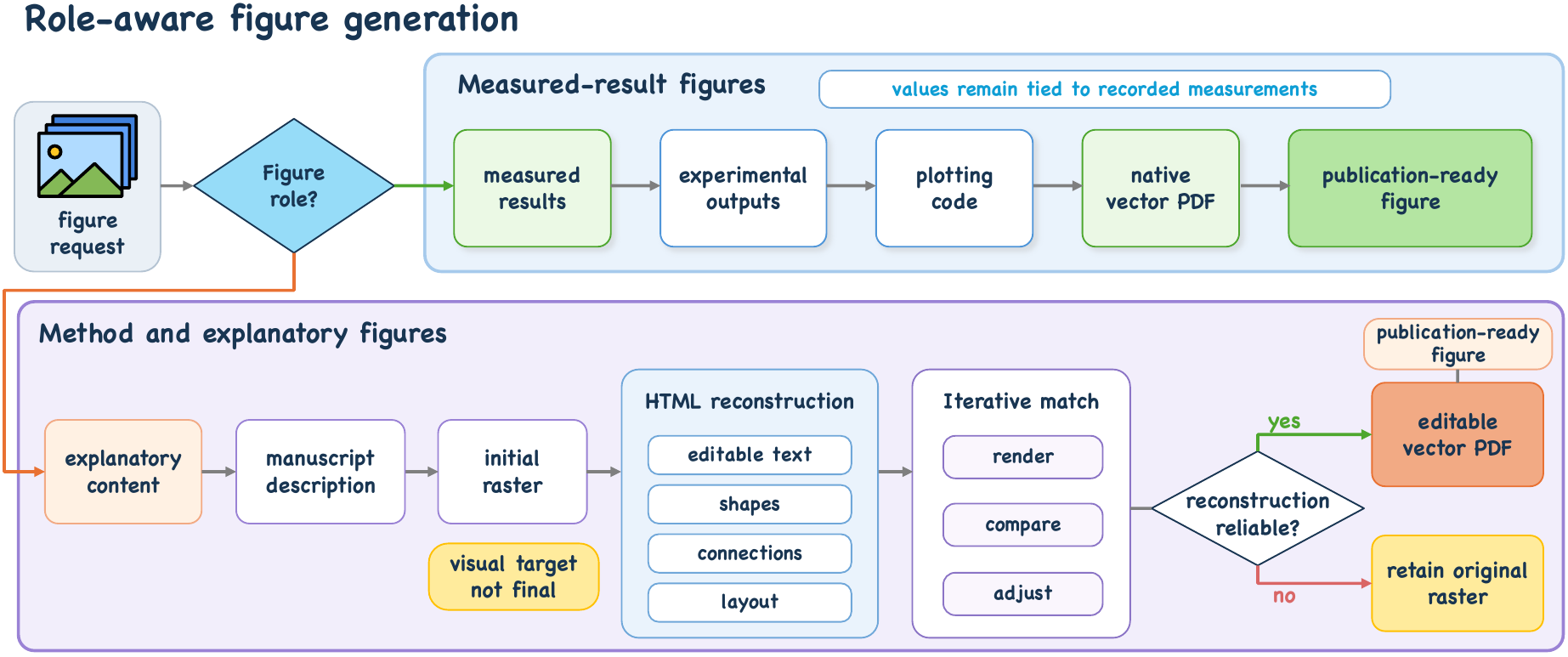}
\vspace{-6mm}
\caption{Role-aware editable figure generation. Measured results follow deterministic plotting to native vector PDF, whereas explanatory content uses a raster visual target followed by iterative HTML reconstruction and vector export; unreliable reconstruction falls back to the raster.}
\label{fig:editable_figure_pipeline}
\vspace{-3mm}
\end{figure}

\subsection{Method and Explanatory Figures}

Method figures, system diagrams, and other explanatory illustrations are generated directly from the manuscript. The image-generation model reads the relevant method description and produces an initial raster figure that captures the intended structure, layout, and visual presentation. We use this generated image as a visual target rather than as the final paper artifact. Spark-to-Paper then uses the coding capabilities of the underlying assistant to reconstruct the figure in HTML, using editable text, shapes, connections, and layout elements. Because the first reconstruction may differ from the generated reference, the system renders the HTML and compares it with the original image. It then iteratively adjusts the layout, geometry, text placement, and visual elements to reduce the discrepancy. In practice, a small number of correction rounds is usually sufficient to recover the main structure and appearance of the generated figure.

Once the reconstruction is accepted, the HTML representation is rendered to PDF. Text and graphical elements created during reconstruction remain editable and vector-based, producing a publication-ready figure without manually redrawing the original image. The generated raster therefore serves primarily as a visual specification, while the reconstructed HTML/PDF provides the final artifact used in the manuscript.

\subsection{Experimental Result Figures}

Figures that report experimental results follow a different path. Because their content corresponds directly to measured values, Spark-to-Paper does not use an image-generation model to create them. Instead, the system reads the experimental outputs and writes plotting code for the required visualization, such as performance comparisons, ablation studies, or trend plots. The plotting program generates the figure directly from the recorded measurements and exports it as a PDF. This keeps the visualized values tied to the underlying experimental results while also producing a vector-format artifact that can be included directly in the manuscript. Thus, generative image models are used for figures whose primary purpose is visual explanation, whereas quantitative figures are generated deterministically from data.

%% file: sections/evaluation.tex
\begin{table}[t]
\centering
\small
\caption{Evaluation dimensions and corresponding measurements used to assess the quality and generation efficiency of Spark-to-Paper.}
\label{tab:evaluation_dimensions}
\renewcommand{\arraystretch}{1.12}
\begin{tabularx}{\linewidth}{
    @{}
    >{\raggedright\arraybackslash}p{0.34\linewidth}
    >{\raggedright\arraybackslash}X
    @{}
}
\toprule
\textbf{Metric} & \textbf{Measurement} \\
\midrule

\multicolumn{2}{@{}l}{\textit{Quality}} \\[2pt]

Citation validity & Resolved references / total references \\
Fabrication detection & Detected unsupported claims / injected claims \\
Figure editability & Editable elements / total figure elements \\
Review precision & Verified issues / raised review issues \\
Cross-template robustness & Successful templates / supported templates \\

\addlinespace[3pt]
\multicolumn{2}{@{}l}{\textit{Efficiency}} \\[2pt]
Generation cost & Monetary cost, tokens, wall-clock time, and deployment footprint \\

\bottomrule
\end{tabularx}
\end{table}

\begin{table}[t]
\centering
\caption{Main comparison of Spark-to-Paper with human-written preprints, prior autonomous research systems, and a single-pass LLM baseline across artifact quality and generation efficiency metrics.}
\label{tab:main_results}
\resizebox{\linewidth}{!}{%
\begin{tabular}{lccccc}
\toprule
System & Citation exist.\ (\%) & Fig.\ editability (\%) & Tokens (M) & USD & Wall-clock \\
\midrule
Human-written preprints (sampled) & 97.8 [94.6, 99.4] & 58 [44, 71] & n/a & n/a & n/a \\
\midrule
 
AI Scientist, released papers \citep{lu2024aiscientist} & 93 (42/45) & 0 (0/210) & n/r & \$10--15 (amort.) & $\sim$12 h / batch \\
AI Scientist-v2, workshop set \citep{yamada2025aiscientistv2} & 91 (58/64) & 3 (0--8) & n/r & $\sim$\$20--25 / attempt & $\leq$15 h / run \\
Agent Laboratory, released paper \citep{schmidgall2025agentlab} & 96 (27/28) & 0 (0/30) & n/r & \$2.33 (gpt-4o) & $\sim$19 min \\
Single-pass LLM draft (same backbone) & 81 (range 76--86) & n/a & 0.11 (0.09--0.13) & \$0.66 (0.55--0.76) & 16 min (13--19) \\
Spark-to-Paper (full stack) & 99.5 [98.4, 100] & 96.4 [92.7, 98.6] & 11.9 [10.2, 13.7] & \$8.1 [6.9, 9.6] & 3.2 h [2.6, 3.9] \\
\bottomrule
\end{tabular}%
}
\vspace{1mm}

\vspace{-4mm}
\end{table}

\begin{table}[t]
\centering
\caption{Ablation study of the Spark-to-Paper quality stack, showing the contribution and incremental cost of gating, self-review, and adversarial review.}
\label{tab:ablation_results}
\resizebox{\linewidth}{!}{%
\begin{tabular}{lcccc}
\toprule
Configuration & Fabr.\ detection (\%) & Review precision (\%) & $\Delta$ tokens (M) & $\Delta$ USD \\
\midrule
Single-pass draft (no gates) & 14 (5/36) [6, 29] & n/a & ref. & ref. \\
Gates only & 69 (25/36) [53, 82] & n/a & +8.1 $\pm$ 0.9 & +5.3 $\pm$ 0.5 \\
Gates + self-review & 81 (29/36) [65, 90] & n/a & +1.1 $\pm$ 0.2 & +0.6 $\pm$ 0.1 \\
Gates + self-review + adversarial review & 92 (33/36) [78, 97] & 74 (42/57) [61, 83] & +2.6 $\pm$ 0.4 & +1.6 $\pm$ 0.2 \\
\bottomrule
\end{tabular}%
}
\vspace{1mm}

\vspace{-5mm}
\end{table}

\section{Evaluation Protocol}
\label{sec:evaluation}

We evaluate Spark-to-Paper using three complementary sources of evidence: controlled experiments, retrospective analysis of existing outputs, and qualitative case studies. Together, these evaluations examine whether Spark-to-Paper can complete the end-to-end research pipeline and produce manuscripts that are reliable, editable, and efficient to generate. Our evaluation focuses on two questions: \textbf{whether Spark-to-Paper produces higher-quality research artifacts than comparable systems, and whether it does so at lower cost}. Whenever possible, comparisons with prior systems are based on their publicly released papers and artifacts, avoiding the need to rerun their pipelines. Our own runs are instrumented directly, while cost figures for prior systems are taken only from their reported results. 

To avoid evaluating a component with the same mechanism used to optimize it, we use independent evaluation procedures whenever possible. For example, citation validity is verified after generation using external bibliographic metadata services rather than the citation checks used inside the pipeline. We also report the provenance of each result and whether optional in-loop checks were enabled during generation.

\subsection{Evaluation Dimensions}

We evaluate Spark-to-Paper along six dimensions, summarized in
Table~\ref{tab:evaluation_dimensions}. The first five measure artifact quality,
while the last measures generation efficiency. For the controlled evaluation, research topics are selected independently of system outputs, and the topic set, model configuration, and evaluation criteria are fixed before generation begins. We register the evaluation protocol with an external timestamp and report all outcomes regardless of whether they favor Spark-to-Paper. The full system is evaluated on eight externally selected topics, three of which are also shared with the single-pass baseline for paired comparison. We report uncertainty across topics rather than relying on individual runs. For prior systems, we use only values explicitly reported in their papers or repositories and mark unavailable measurements as \textit{not reported} rather than estimating them.

\subsection{Main Results}

Table~\ref{tab:main_results} compares Spark-to-Paper with human-written preprints, previously released autonomous research systems, and a single-pass LLM baseline. Spark-to-Paper achieves the highest citation validity among the evaluated systems, with 99.5\% of references successfully resolved, while preserving 96.4\% of figure elements in editable form. In comparison, the released papers from prior autonomous research systems exhibit citation validity between 91\% and 96\%, but provide little or no figure editability. The single-pass baseline is substantially cheaper and faster, but its citation validity drops to 81\%, illustrating the quality-efficiency trade-off introduced by the full pipeline.

The comparison combines controlled measurements with retrospective audits. Results for Spark-to-Paper and the single-pass baseline are obtained from our instrumented runs on pre-registered topics. The full system is evaluated on eight topics, while three of these topics are additionally run with the single-pass baseline for paired comparison. Human-written preprints serve as a reference point for artifact quality and are evaluated using the same external citation and figure-analysis procedures. By contrast, results for AI Scientist, AI Scientist-v2, and Agent Laboratory are obtained by auditing artifacts publicly released by the corresponding systems; we do not rerun these systems or normalize them to our model backbone, topics, or pricing environment.

For our measurements, citation validity is computed over 384 references from eight Spark-to-Paper campaign papers; the human-preprint reference comprises 320 references from eight sampled papers. Figure editability for Spark-to-Paper is evaluated over approximately 1{,}900 ground-truth figure elements, excluding figures that are intentionally rasterized by design. Reported intervals are cluster-bootstrap intervals over papers; when only three paired runs are available, we report the observed range instead of a confidence interval. Cost and runtime for our systems are measured directly from execution logs. For prior systems, we report only values explicitly stated in their papers or repositories, using the pricing assumptions of their original publications rather than retrospectively repricing them.   We use \textit{n/a} for structurally inapplicable quantities and \textit{n/r} for quantities not reported by the original system.

\begin{figure}[t!]
\centering
\includegraphics[width=0.98\linewidth]{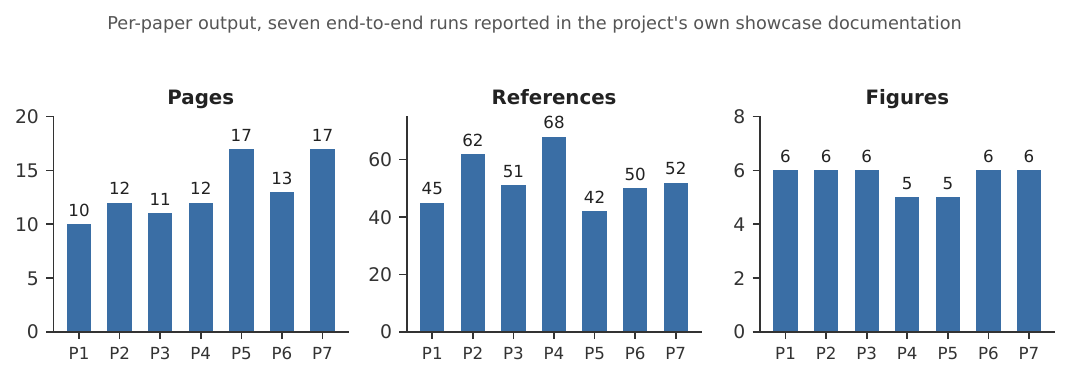}
\vspace{-3mm}
\caption{Page, reference, and figure counts for the end-to-end papers documented in the project's own showcase materials. Self-reported by the system's maintainers, not an independent measurement; included as existence evidence, not as a benchmark result.}
\label{fig:showcase}
\vspace{-3mm}
\end{figure}

\subsection{Ablation Study}

Table~\ref{tab:ablation_results} isolates the contribution of the major components in Spark-to-Paper's quality-control stack. Starting from a single-pass draft, adding deterministic and model-based gates increases fabrication detection from 14\% to 69\%. Adding self-review further raises detection to 81\%, while the complete stack with adversarial review reaches 92\%. The adversarial-review stage also achieves 74\% review precision, indicating that most issues raised by the reviewer correspond to verifiable problems rather than spurious critiques. These gains come with additional inference cost: relative to the preceding configuration, gates add the largest token and monetary overhead, while self-review and adversarial review provide further quality improvements at smaller incremental cost.

We evaluate fabrication detection using a fixed corpus of 36 seeded probes spanning ten failure families and drawn from three sources. All probes are injected into the shared source material before drafting at the same intervention point, and the identical probe set is used for all four configurations. The detection procedure is also held fixed across configurations. Consequently, differences between rows reflect changes in the quality-control stack rather than changes in either the injected failures or the evaluation instrument. Detection intervals are Wilson 95\% confidence intervals over the 36 probes.

Review precision is evaluated separately because it is defined only for configurations containing an explicit review stage. We sample 60 issues raised during review and have them assessed by blinded raters. Three ``cannot tell'' judgments are excluded, leaving 57 issues in the reported denominator. The resulting precision therefore measures the fraction of review findings that correspond to independently verifiable issues.

The final two columns quantify the incremental computational cost of each added layer. Token and dollar deltas are measured on the three topics shared across configurations and are reported relative to the immediately preceding row. These increments are therefore intended to show the marginal cost of each quality-control component, rather than to reconstruct the eight-topic average reported in Table~\ref{tab:main_results}. Because the ablation uses the three-topic paired subset, its cumulative cost does not necessarily equal the full-system mean in the main comparison. We use \textit{n/a} when a metric is undefined for a configuration and \textit{ref.} for the single-pass reference configuration.

 \begin{figure}[t!]
    \centering
    \includegraphics[width=\linewidth]{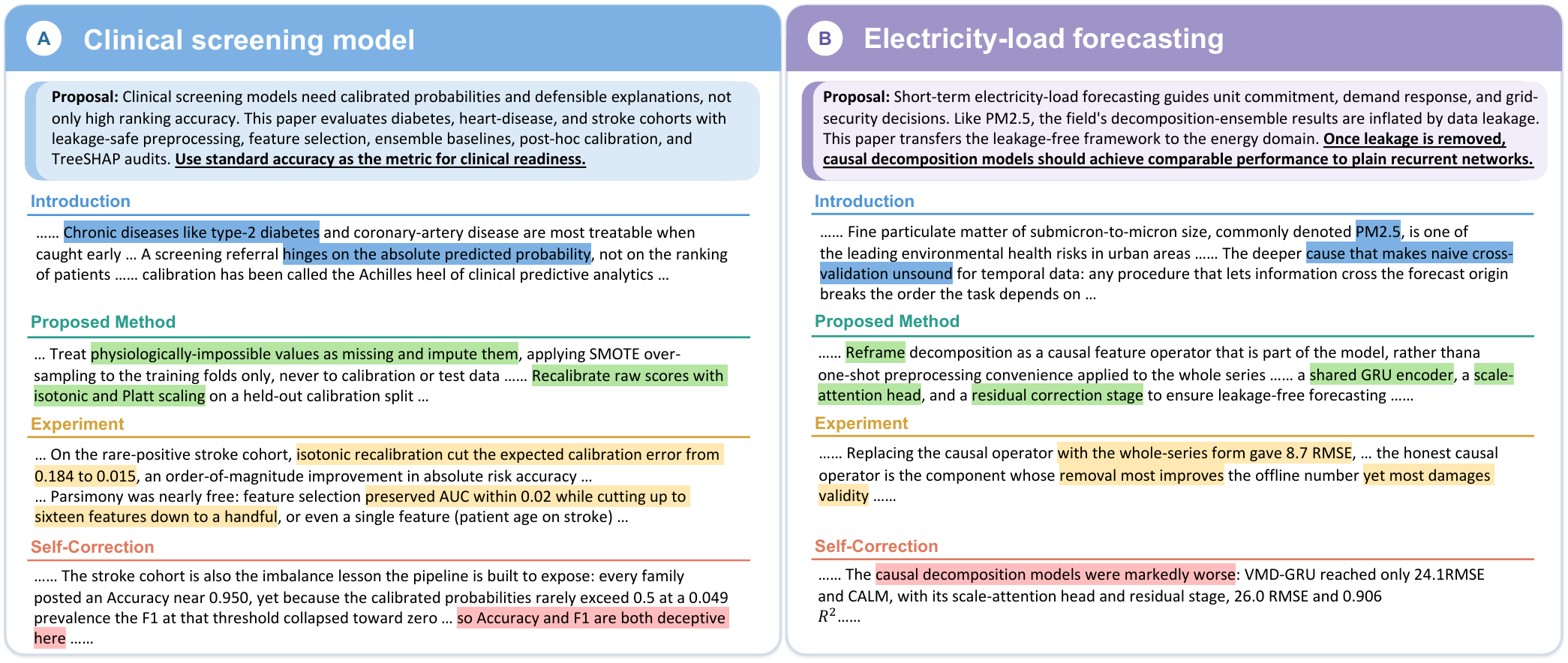}
    \vspace{-5mm}
    \caption{Case study on two different domain demo papers with only one short proposal as inputs, incorrect expectations are highlighted by Bold and Underline text.}
    \label{fig:case_study}
    \vspace{-3mm}
\end{figure}

\subsection{Case Study}

We show two case studies in Fig. \ref{fig:case_study} that demonstrate Spark-to-Paper's ability to autonomously identify research gaps, synthesize novel methods, and maintain academic integrity. The generated answers also criticize the drawbacks of current approaches and highlight the drawbacks and research gaps in a specific domain. In clinical risk screening, the framework goes beyond raw discrimination accuracy to address critical miscalibration and interpretability, introducing a leakage-safe pipeline that integrates post-hoc probability recalibration with TreeSHAP driver audits. In time-series forecasting, Spark-to-Paper redefines traditional decomposition by framing it as a causal, sliding-window feature operator to eliminate look-ahead bias. Crucially, by conducting an empirical leakage audit, the system fearlessly exposes how conventional whole-series decomposition inflates accuracy by borrowing information from the future, demonstrating its commitment to evidence-based scientific truth.

Furthermore, we inject two incorrect expectations into the proposal to demonstrate the effectiveness of our model's refinement and revision capabilities. In the clinical screening example, the proposal mistakenly positions Accuracy as the primary evaluation metric, a common pitfall in the medical domain due to severe data imbalance. As reflected by the paper's results, the proposed method implements multiple evaluation metrics to cross-validate the trustworthiness of the final results, ultimately concluding that both Accuracy and F1 scores are highly deceptive. In the PM2.5 forecasting domain, we inject another strong prior expectation from the user (that causal models would achieve comparable performance). The paper's actual empirical results present a verifiable conflict with this claim, showing that decomposition models perform markedly worse. This scenario effectively tests the model's ability to prioritize objective experimental evidence over misleading user prompts.

%% file: sections/conclusion.tex
\section{Conclusion}
We present Spark-to-Paper, an end-to-end research paper generation system that consists of thirteen composable skills inside an existing coding assistant. Spark-to-Paper reuses the assistant's native capabilities for file interaction, tool use, and code execution, while separating model-based judgment from deterministic operations that support direct verification. Beyond manuscript drafting, the system plans experiments around the evidence required by the paper, revises claims according to empirical results, combines deterministic integrity checks with long-horizon self-critique, and produces editable vector figures through programmatic plotting or code-based reconstruction. Our experiments show that this lightweight design produces complete research artifacts with high citation validity and figure editability while substantially improving the detection of unsupported claims. Importantly, Spark-to-Paper does not force every research trajectory to succeed: when repeated experimentation fails to support the original hypothesis, the system bounds the resulting Self-Refutation Loop, preserves the unsuccessful trajectory as a research report, and moves on to a new idea. This design makes automated research a traceable process in which the system revises or abandons claims according to evidence rather than forcing them toward a successful narrative.

%% file: sections/appendix.tex
\newpage

\section{Deterministic Gates and Failure Conditions}
\label{app:deterministic_gates}

Spark-to-Paper uses deterministic gates for properties with explicit correctness
criteria. Each gate reads the artifacts produced by the current stage, emits a
structured report, and returns a pass or fail decision before those artifacts
are passed downstream. We distinguish between \emph{fatal} violations, which
stop execution, and \emph{warnings}, which are recorded but do not block
progress. Fatal conditions are reserved for violations that make an artifact
invalid, inconsistent, or unverifiable.

The \textbf{Template Gate} and \textbf{Blueprint Gate} enforce structural
consistency before manuscript generation. The former verifies that the venue
specification contains the required document structure, section constraints,
citation settings, template resources, and assembly information. The latter
checks that the planned manuscript conforms to these requirements, including
its sections, contributions, figures, tables, citation types, and title
constraints. Violations with an unambiguous correction may be normalized
automatically; inconsistencies that require semantic judgment cause the gate to
fail. 

The \textbf{Citation Gate} checks bibliographic integrity, including malformed
or duplicate entries, unresolved citation keys, unused references, and
inconsistencies between citations and their associated claims. When identifier
resolution is enabled, DOI, URL, or preprint records are additionally verified.
Deterministically invalid identifiers cause failure, whereas temporary network
or service errors are reported as warnings. The \textbf{Manuscript Gate}
checks structural requirements together with result integrity. It detects
unresolved placeholders, invalid result-table structure, missing notation
definitions, and related venue-specific constraints. Result-integrity checks
follow the execution mode selected during input routing: unobserved empirical
results are rejected in Proposal Mode, while quantitative statements that
cannot be grounded in the supplied evidence are flagged in Data-Aware Mode.
If the evidence required for verification is itself unavailable, the gate
fails rather than relying on model judgment. 

The \textbf{Figure Gate} verifies that required figure artifacts exist and that
their representation and generation path are consistent with their intended
role. Quantitative figures must remain grounded in measured results, while
explanatory figures claimed to be editable must retain the corresponding
editable structure. Raster output remains permissible when the content is
inherently raster or when reliable reconstruction is unavailable. Finally, the
\textbf{Compilation Gate} requires the assembled \LaTeX{} project to compile
successfully with citations and cross-references resolved. Compilation repair is
bounded; unresolved failures are reported rather than triggering an indefinite
repair loop. 

These gates deliberately cover only properties that can be verified without
semantic judgment. They can determine whether a required artifact exists,
whether a reported value is grounded in available evidence, whether a citation
resolves, or whether a project compiles. They do not determine whether a
contribution is important, an argument is persuasive, an experimental design is
scientifically appropriate, or a figure communicates its message effectively.
Such questions are handled by the model-based review and claim-assessment
mechanisms described elsewhere in the paper. Deterministic gating therefore
does not attempt to automate every quality decision; instead, it makes directly
verifiable requirements non-negotiable once they have been specified.

\section{Experiment Execution Details}
\label{app:experiment_execution}

Spark-to-Paper does not directly write experimental outputs into the manuscript.
Instead, experiment execution is treated as an evidence-grounded process. The
system first identifies the evidence required by the current manuscript claims,
determines which corresponding experiments are feasible, and executes those
experiments using the available code and data. The resulting measurements are
then verified before they can be treated as experimental evidence.
Algorithm~\ref{alg:experiment_execution} summarizes this procedure.

The feasibility decision preserves the experimental design committed to during
the planning stage. Spark-to-Paper does not replace a specified evaluation with
an easier proxy, a different data construction, or synthetic observations when
doing so would change the experiment described by the manuscript. When a
required resource is unavailable, the system records the missing dependency and
leaves the corresponding result unspecified rather than substituting a
generated value. This prevents the evaluation protocol from being adapted to
what is easiest to execute after the intended evidence has already been
specified.

For feasible experiments, the system records the configuration, dataset,
random seed, metric definition, logs, and raw measurements produced during
execution. A measurement is accepted as verified experimental evidence only
when it can be traced to these source artifacts and, for aggregated results,
recomputed from the underlying runs. The verification step also checks that
reported experiments were actually executed, that the implementation contains
the components, baselines, and ablations described by the manuscript, and that
the evaluation protocol does not contain detectable failures such as data
leakage, invalid splits, or mismatched metrics. Measurements that fail these
checks are rejected as experimental evidence rather than completed or repaired
through generation.

\begin{algorithm}[!t]
\caption{Evidence-Grounded Experiment Execution}
\label{alg:experiment_execution}
\begin{algorithmic}[1]
\Require Manuscript $M$, experimental design $D$, available code $C$, data $X$
\Ensure Revised manuscript $M'$, experiment artifacts $A$
\State Identify claims $\mathcal{Q}$ in $M$ that require experimental evidence
\State Derive required experiments $\mathcal{E}$ from $D$ and $\mathcal{Q}$

\For{each experiment $e \in \mathcal{E}$}
    \State Assess whether $e$ is feasible using $C$ and $X$

    \If{$e$ is feasible}
        \State Execute $e$ and record configurations, seeds, logs, and measurements
        \State Verify result provenance and experimental consistency

        \If{the result passes verification}
            \State Admit the measured result into $A$
            \State Update the corresponding tables and figures
        \Else
            \State Reject the result and revise the associated claim
        \EndIf

    \Else
        \State Record the missing resource
        \State Keep the corresponding result unspecified
    \EndIf
\EndFor

\State Re-evaluate manuscript claims using the admitted evidence
\State Propagate revisions across evidence-dependent sections
\State Preserve negative, null, and inconclusive results
\State Recompile and validate the revised manuscript

\If{the central research objective remains unsupported}
    \State Pass the trajectory to the bounded Self-Refutation procedure
\EndIf
\State \Return $M', A$
\end{algorithmic}
\end{algorithm}

\section{Claim Admission Protocol}
\label{app:claim_admission}

Given the verified evidence produced by experiment execution, Spark-to-Paper
re-evaluates manuscript claims rather than assuming that the initial narrative
remains valid. Before claim-level assessment, the system identifies the primary
contribution type of the paper, such as a framework, model, benchmark, dataset,
system, or empirical study. This distinction matters because different claims
require different forms of evidence. For example, failure of one model instance
to outperform a baseline may weaken a model-level performance claim without
necessarily invalidating a broader framework contribution. Claims are therefore
assessed relative to the contribution they are intended to support rather than
against a single result table.

Each major claim is assigned one of five evidence labels:
\textit{supported}, \textit{partially-supported}, \textit{unsupported},
\textit{contradicted}, or \textit{needs-confirmation}. A claim can be labeled
\textit{supported} or \textit{partially-supported} only when it is linked to
specific verified evidence. The assigned label determines the corresponding
revision action, as summarized in Table~\ref{tab:claim_admission}.

\begin{table}[t]
\centering
\small
\caption{Claim admission labels and corresponding revision actions.}
\label{tab:claim_admission}
\begin{tabularx}{\linewidth}{
    @{}
    >{\raggedright\arraybackslash}p{0.4\linewidth}
    >{\raggedright\arraybackslash}X
    @{}
}
\toprule
Label & Action \\
\midrule
\textit{supported} &
Retain with evidence-matched wording \\
\textit{partially-supported} &
Narrow the claim or request additional evidence \\
\textit{unsupported} &
Run a feasible missing experiment, weaken, or remove \\
\textit{contradicted} &
Remove or report as a limitation \\
\textit{needs-confirmation} &
Return the unresolved claim for author confirmation \\
\bottomrule
\end{tabularx}
\end{table}

\begin{table}[t]
\centering
\small
\caption{Main persistent artifacts used by Spark-to-Paper.}
\label{tab:project_artifacts}
\begin{tabularx}{\linewidth}{
    @{}
    >{\raggedright\arraybackslash}p{0.3\linewidth}
    >{\raggedright\arraybackslash}p{0.2\linewidth}
    >{\raggedright\arraybackslash}X
    @{}
}
\toprule
Artifact & Stage & Purpose \\
\midrule
\texttt{blueprint.json} & Planning & Paper structure, claims, notation, experiments \\
\texttt{template.json} & Planning & Venue specification and execution mode \\
\texttt{refs.bib} & Citation & Verified bibliography \\
\texttt{claims\_map.json} & Citation & Claim--citation associations \\
\texttt{sections/*.tex} & Writing & Section-level manuscript sources \\
\texttt{figures/} & Figure & Figures and generation records \\
\texttt{results.facts.json} & Data & Grounded quantitative evidence \\
\texttt{main.tex/pdf} & Assembly & Final manuscript project \\
\texttt{logs/*.io.md} & All stages & Stage-level input and output records \\
\bottomrule
\end{tabularx}
\vspace{-4mm}
\end{table}

Claim weakening changes the scope of a statement rather than merely reducing
its rhetorical strength. Claims of statistical significance, robustness,
generalization, or broad empirical superiority are retained only when the
corresponding evidence has been obtained. When the evidence supports a narrower
conclusion, the manuscript is revised to state that narrower conclusion
explicitly. Negative, null, and inconclusive results are preserved rather than
omitted, and the overall research narrative is allowed to weaken when required
by the measured evidence. 

Revisions are propagated across all occurrences of the affected claim. A claim
that appears in the abstract, introduction, results, and conclusion is updated
consistently rather than corrected only where the corresponding experiment is
reported. The abstract is revised last so that it reflects the final claim
strength of the completed manuscript. Claims marked
\textit{needs-confirmation} cannot remain unresolved in the final deliverable;
they must be supported, revised, removed, or explicitly resolved by the author.

The protocol separates machine-checkable evidence integrity from semantic claim
assessment. Deterministic programs verify the integrity of the experimental
evidence and the final manuscript, while determining whether a particular piece
of evidence semantically supports a claim is currently performed by the model.
Each assessment is recorded in a structured report containing the original
claim, its manuscript location, assigned label, supporting evidence, revision
action, and rationale. Claim admission therefore remains model-based where
semantic judgment is required, but the resulting decisions are explicit and
auditable rather than being hidden inside manuscript rewriting.

\section{Implementation Details}
\label{app:implementation}

We provide additional implementation details on input routing, persistent
project artifacts, and venue-specific configuration below.

The pipeline first routes the user input according to its current research
state. A short idea may be expanded into a structured proposal, whereas a
developed proposal can enter the paper-generation pipeline directly. The system
also determines whether measured experimental evidence is already available and
selects either Proposal Mode or Data-Aware Mode. This mode is stored in the
project state and propagated to downstream stages. In Proposal Mode, unavailable
empirical results remain unspecified; in Data-Aware Mode, quantitative
statements must be grounded in the supplied data or experimental outputs. The
same downstream pipeline can therefore operate both before and after experiments
have been conducted.

Each stage materializes its output as an explicit project artifact. Planning
produces the paper blueprint and active template specification; citation
retrieval produces a verified bibliography and claim--citation mapping; writing
produces section-level \LaTeX{} sources; figure generation writes visual
artifacts and their associated records; and assembly produces the complete
\LaTeX{} project and compiled manuscript. When measured results are available,
their extracted quantitative facts are stored separately so that downstream
integrity checks can distinguish observed evidence from generated text.
Table~\ref{tab:project_artifacts} summarizes the main persistent artifacts used
for communication between stages.

Venue-specific behavior is parameterized through the template specification
rather than hard-coded into individual skills. The specification defines the
section structure, length constraints, citation style, figure and table
requirements, title and keyword constraints, and assembly conventions consumed
by downstream stages. Supporting a new venue therefore requires a new
specification and the corresponding template resources rather than changes to
the pipeline logic. Template provenance is retained explicitly: official or
user-provided styles can be used for submission, while bundled approximations
are marked as demonstration-only and must be replaced before submission.

\section{Generated Paper Examples}

We additionally provide several examples of papers generated by Spark-to-Paper.
The following figures show representative sections from complete end-to-end
outputs, including the introduction, method, experimental results, and analysis.
These examples are included to illustrate the structure, level of detail, and
visual form of the generated manuscripts, rather than as an additional
quantitative evaluation.

\newpage

\begin{figure}[htbp]
    \centering
    \includegraphics[width=1\linewidth]{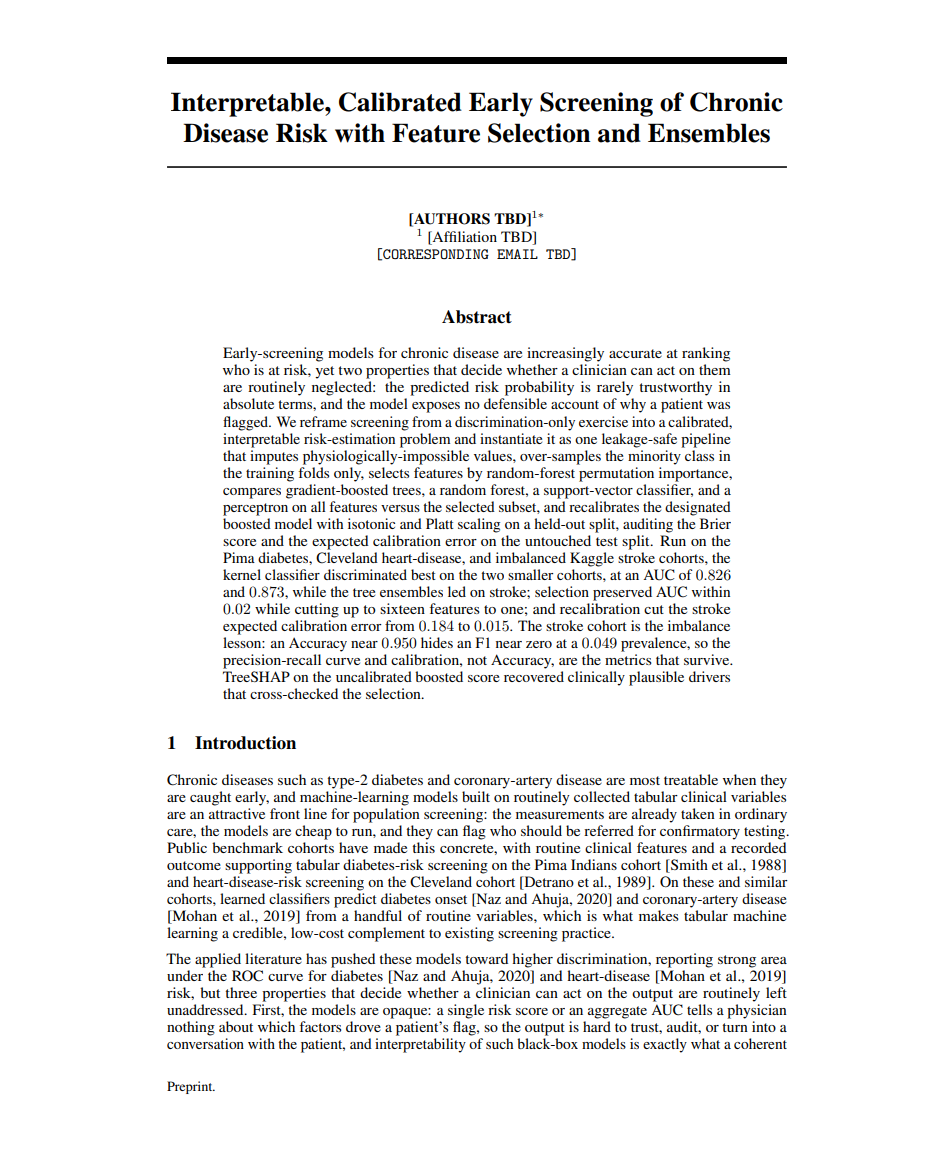}
    \caption{Paper Demo (Part 1): Screenshot for the demo paper introduction.}
    \label{fig:paper_demo_1}
\end{figure}

\begin{figure}[htbp]
    \centering
    \includegraphics[width=\linewidth]{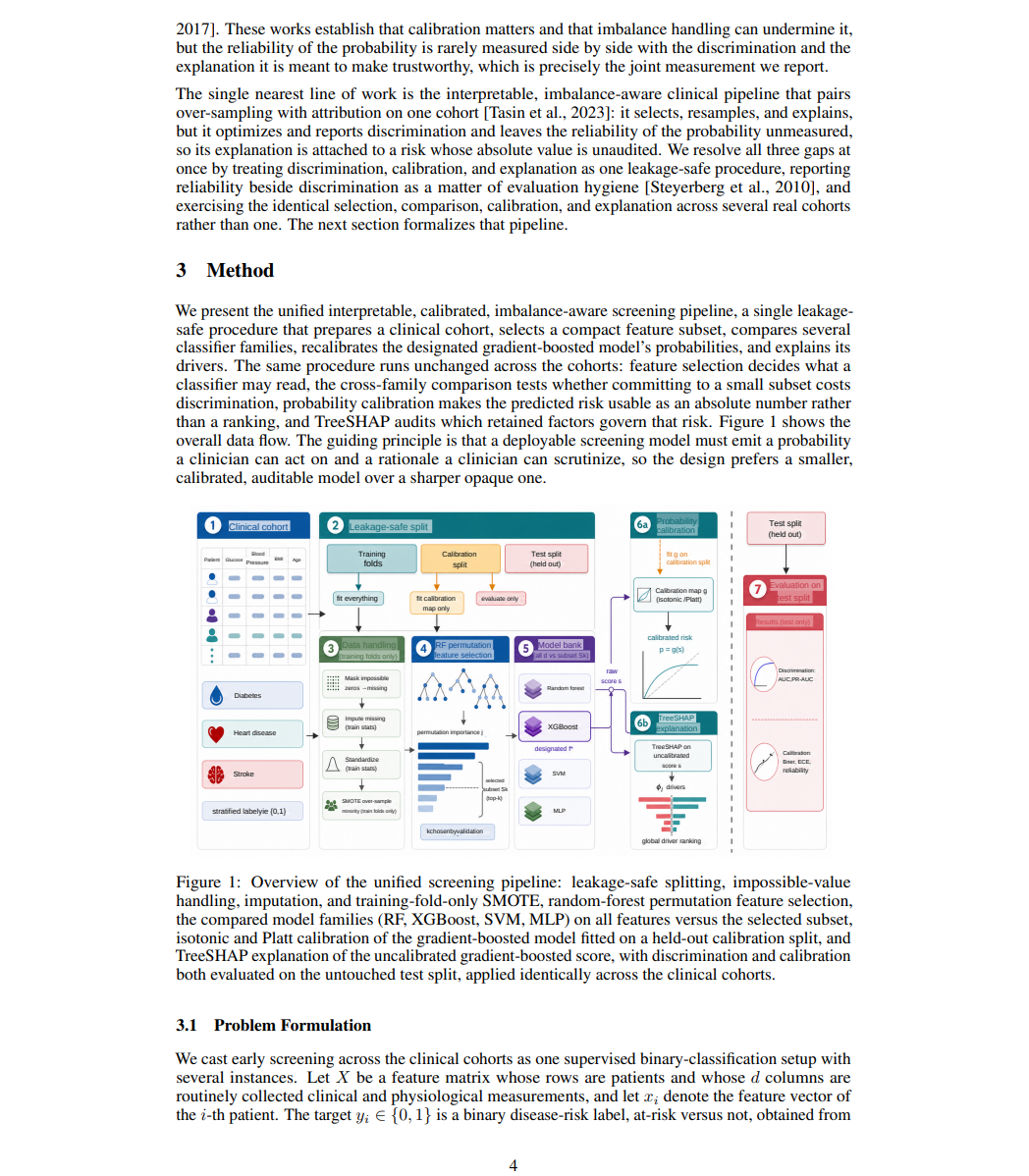}
    \caption{Paper Demo (Part 2): Screenshot for the demo paper method.}
    \label{fig:paper_demo_2}
\end{figure}

\begin{figure}[htbp]
    \centering
    \includegraphics[width=\linewidth]{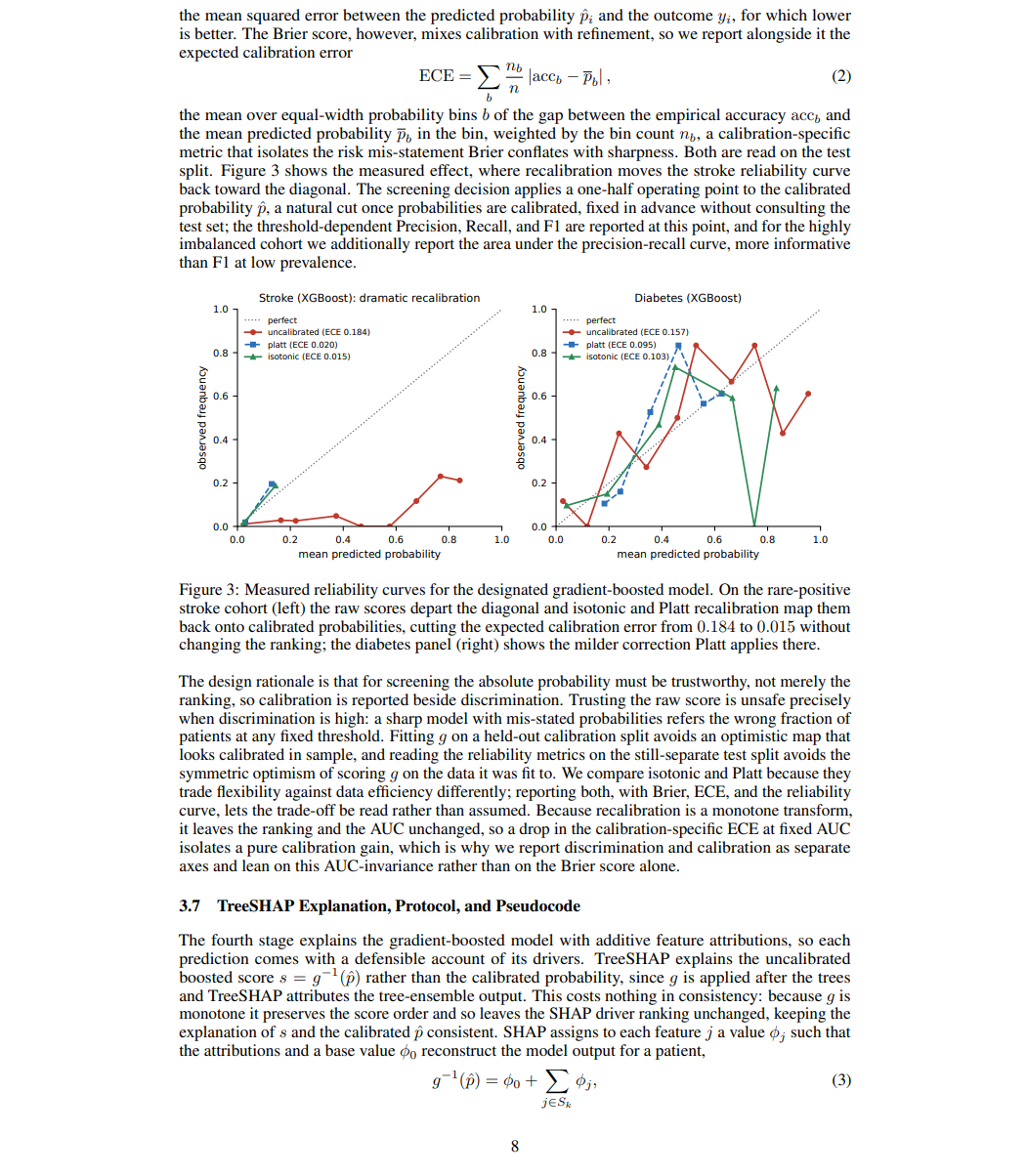}
    \caption{Paper Demo (Part 3): Screenshot for the demo paper experiment and analysis.}
    \label{fig:paper_demo_3}
\end{figure}

\begin{figure}[htbp]
    \centering
    \includegraphics[width=\linewidth]{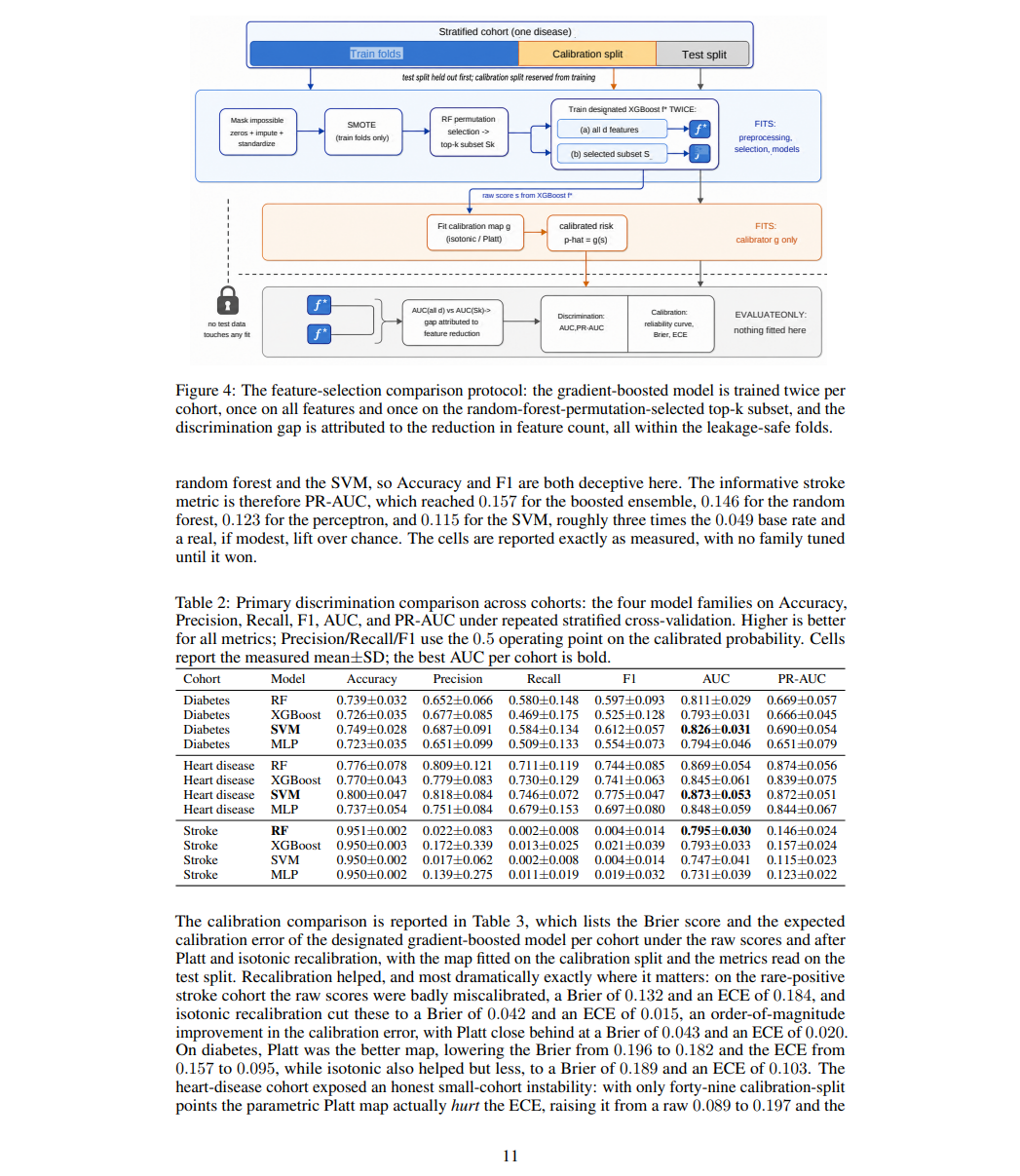}
    \caption{Paper Demo (Part 4): Screenshot for the demo paper experiment and analysis.}
    \label{fig:paper_demo_4}
\end{figure}